\UseRawInputEncoding
\documentclass[
]{ceurart}

\begin{document}

\title{Ontomathedu Ontology Enrichment Method}

\author[1]{Konstantin S. Nikolaev}[%
orcid=0000-0003-3204-238X,
email=konnikolaeff@yandex.ru,
]
\address[1]{Federal State Institution of the Federal Research Center NIISI RAS}

\address[2]{Kazan Federal University}

\author[2]{Olga A. Nevzorova}[%
orcid=0000-0001-8116-9446,
email=onevzoro@gmail.com,
]

\begin{abstract}
  Nowadays, distance learning technologies have become very popular. The recent pandemic has had a particularly strong impact on the development of distance education technologies. Kazan Federal University has a distance learning system based on LMS Moodle. This article describes the structure of the OntoMathEdu ecosystem aimed at improving the process of teaching school mathematics courses, and also provides a method for improving the OntoMathEdu ontology structure based on identifying new connections between contextually related concepts.

\end{abstract}


\maketitle

\section{Introduction}
The ontological approach is often used to organize training in a more structured format, as well as to help the student form a full picture of the current discipline, to more clearly understand the existing patterns and relationships between concepts in the problem area under consideration. The basis of such an educational infrastructure can serve as a qualitative ontology. Currently, there are many training systems that use various ontologies to organize subject data. For example, Lutoshkina et al. in the work \cite{Lutoshkina2011}, the ontology of the subject area is used to systematize the multimedia content of electronic courses (the connections between the concepts of ontology are used to construct an integral structure of the lesson). In addition, Stancin et al. \cite{Stancin2020} provide a detailed overview of the use of ontologies in education, among which the most interesting for us are ontologies describing the subject under study and e-learning services. Here are some examples of such solutions:

1. LiReWiapproach (Conde et al. \cite{Conde2019}). Educational ontology is used to describe the topics to be mastered by students and the pedagogical relations between the topics.

2. MOOC-KG (Dang et al. \cite{Dang2019}). To build a knowledge graph, an ontology is used that models knowledge about online learning resources.

3. MKMSE (Martinez-Ramirez et al. \cite{Martinez-Ramirez2018}). Ontology is used to store mathematical knowledge.

4. Ontology is used to represent the semantic description of document resources and the relationships between document resources and other types of resources (Hai \cite{Hai2019}).

5. CALMS (Erazo-Garzon et al. \cite{Erazo-Garzon2019}). Ontologies are used to define concepts and semantic relations between academic information and conceptual aspects.

6. Ontology is used as an expression of knowledge to support the ends of the content of education. (Kubekov et al. \cite{Kubekov2018})

\section{Structure of the OntoMathEdu ecosystem}
\begin{figure}
  \centering
  \includegraphics[width=\linewidth]{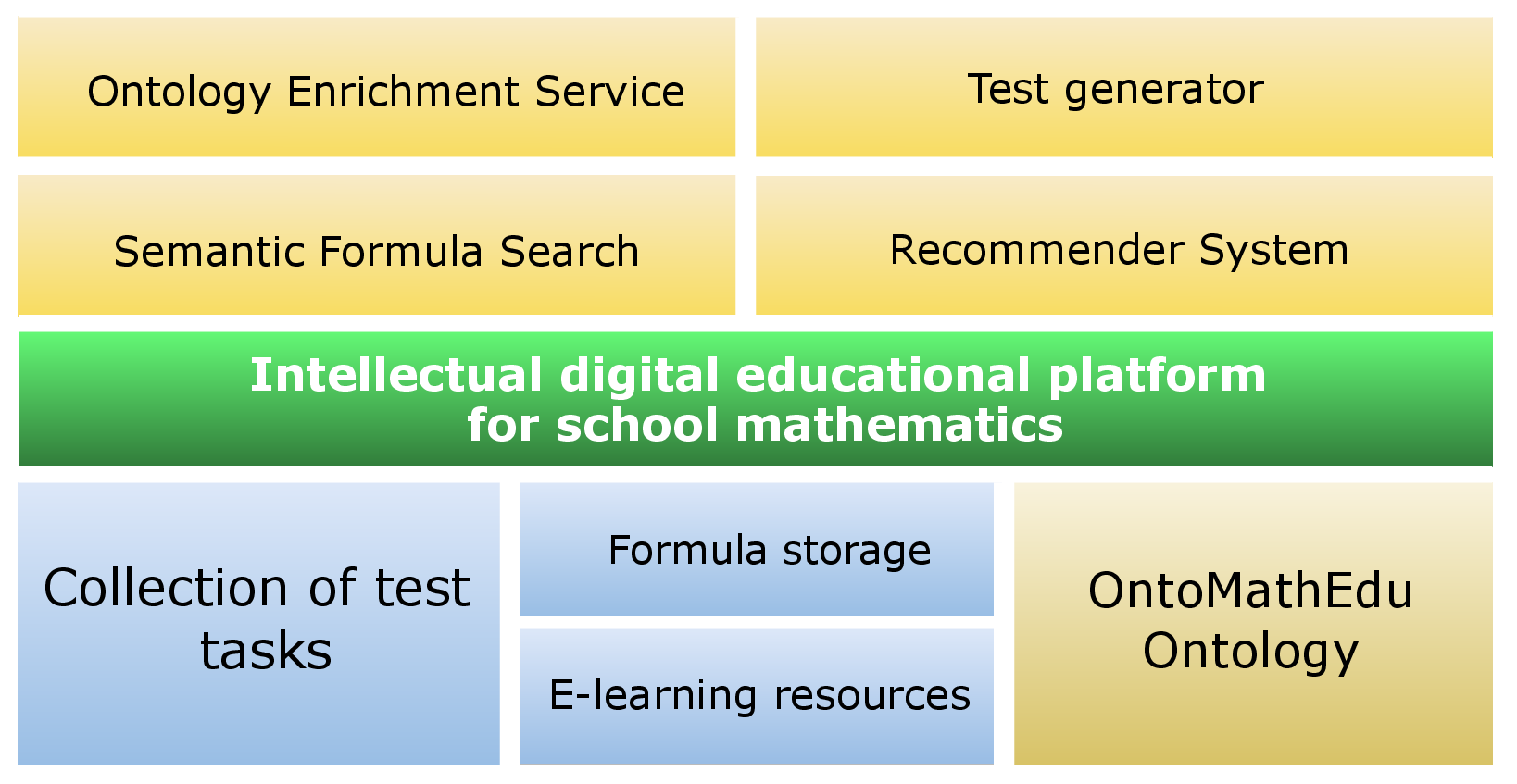}
  \caption{Components of the OntoMathEdu ecosystem}
  \label{fig:ecosystem}
\end{figure}

The article discusses the structure and main tasks of the infrastructure of software components built around the ontology of school mathematics OntoMathEdu \cite{Kirillovich2020, Kirillovich2019}. Ontology OntoMathEdu is a knowledge system that describes a set of concepts, relationships between them and statements within the framework of a school course in planimetry. Figure \ref{fig:ecosystem} shows the composition of the OntoMathEdu digital ecosystem. The ecosystem consists of the following components.

\begin{enumerate}
\item Intelligent digital educational platform for school mathematics. This component occupies a central place in the OntoMathEdu ecosystem. Its main purpose is the application of semantic and ontological technologies in teaching school mathematics.
\item Collection of questions. This collection is used as an input data set for various components of the ecosystem, in particular, for the module "Ontology Enrichment Services", as well as in the central component "Intelligent Digital Educational Platform for School Mathematics". The collection is constantly updated.
\item Formula storage. This component contains formulas extracted from school geometry textbooks. Formulas are presented in various formats (plain text, LaTeX, OpenMath). We plan to use this repository as an auxiliary data set for building test tasks and other information components.
\item Digital educational resources. This component combines all auxiliary data sources located on the Internet (for example, open educational databases, resources for the construction of geometric shapes, etc.)
\item OntoMathEdu ontology. The OntoMathEdu ontology is a reflection of the level of knowledge corresponding to the level of school mathematics. It serves as the main repository of concepts and their relationships that are used by other services.
\item Ontology enrichment service. This component includes a set of methods that are used to clarify the relationships between the concepts of ontologies and improve the horizontal connectivity of ontologies.
\item Test generator. This component is used to automatically create new test tasks based on the analysis of the structure and concepts of existing tasks.
\item Search by semantic formulas. A search module that performs semantic search of mathematical texts present in the ecosystem. As a result of the search, it returns a set of texts containing the requested concept and offers a list of related concepts for further navigation through the text database.
\item Recommendation system. This module helps the user to study various sections of the training course, providing an individual learning trajectory.
\end{enumerate}

\section{Ontology enrichment service}
The ontology OntoMathEdu is currently in the active stage, and contains more than 900 concepts and more than 20 relationships. The task of introducing new relations between concepts is urgent. We have developed a method for introducing new relations between concepts based on the use of a collection of questions. To find new connections between concepts, we use a collection of test questions on school geometry. Our main idea is as follows. Each question contains some basic concepts of ontology, about which the question is asked. We assume that the answer to this question can be found with the help of ontology, in which there is a chain of concepts of limited length connecting the original concepts. In addition, the search for such connections allows us to evaluate the structural properties of the ontology, namely the horizontal connectivity of the resource.

The source of the texts is a collection of school test tasks on geometry with the markup of nominal groups. These nominal groups will act as candidates to search for the corresponding concept in the OntoMathEdu ontology. A noun group of the first type (Noun phrase, NP) is understood as a syntactic construction in which there is a vertex (noun) and dependent words. An example of NP is a "finite set of points", where the real "set" is the vertex of a given phrase with two dependent words. A nominal group of the second type (Prepositional phrase, PP) is a phrase in which the main word is a preposition. An example of PP is the phrase "up to two characters after the dot", where the preposition "to" is the vertex of the nominal group.

Then, for each pair of concepts, all existing relationships are searched, and the search is performed twice: in the first case, only hierarchical relationships (rdfs:subClassOf and ome:hasChild) are used as relationships between concepts, and in the second case, all possible relationships are used. We call the first type of relationship as a "hierarchical relationship", and the second as a "complete relationship". The presence of an optimal relationship is an event in which the shortest complete link is shorter than the hierarchical one. This connection is optimal, because any pair of concepts is somehow connected by a hierarchical connection (we can always climb from the first concept to the node shared with the second concept and descend to the second concept), but this connection will not be the shortest if there are direct connections between the concepts under consideration.

The resulting set of shortest links with an indication of optimality is sent for analysis to a group of experts for manual updating of the ontology. It can also be used as auxiliary information when checking test tasks.
\begin{figure}
  \centering
  \includegraphics[width=\linewidth]{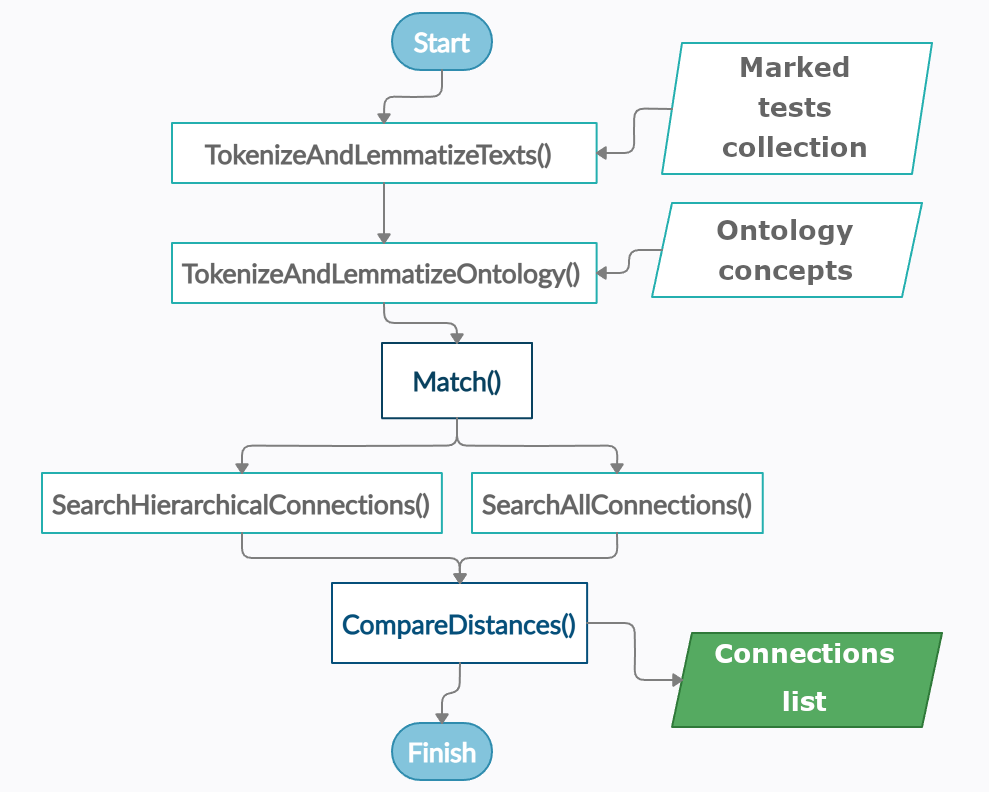}
  \caption{Block diagram of the ontology enrichment method}
  \label{fig:enrich}
\end{figure}
Figure \ref{fig:enrich} shows a block diagram of the task of enriching ontology through method names. The following sections will describe the main stages of the algorithm for finding optimal connections between concepts in the texts of questions with the name of the method.
Here are some comments on Figure \ref{fig:enrich}.
\begin{enumerate}
\item Extraction of NP and PP. At this stage, we need to select a set of NP and PP in the texts of questions and text variants of answers (this means that we do not consider answer options for computational problems and for problems with the choice of an answer presented as a number or a set of characters that are not a dictionary unit). Thanks to the already preliminary markup of the text by the nominal groups NP (TERM1 tag) and PP (TERM2 tag), this stage is reduced to highlighting the contents of these tags using the standard xml library for Python 3.

\item Tokenization and normalization of phrases from nominal groups (NP and PP): TokenizeAndLemmatizeTexts() function. Tokenization of phrases received at the last stage is performed using the nltk library. Normalization of the resulting set of tokens is carried out using the pymorphy2 library, which has the necessary functionality to bring words to a normal form. For example, for the word "triangles", the normal form is "triangle".

\item Tokenization and normalization of ontology concepts: TokenizeAndLemmatizeOntology() function. Tokenization of phrases received at the last stage is performed using the nltk library. Tokens are normalized using the pymorphy2 library.

\item Match search: Match() function. Next, we need to find similar sequences of tokens in the question text and names of ontological concepts, using a modified Jacquard measure to work with a sequence of strings. The sequence element in our case is a lemma from the phrases NP or PP. However, since the words in the compared sequences may not completely match, we also apply the Jacquard measure when searching for common words between sequences. In this case, the elements of the sequence will be individual letters.

When performing this method, we will introduce two thresholds: a lower accuracy threshold at which two words are considered the same, and a lower accuracy threshold at which two sequences of words are considered similar. Their values were obtained experimentally on a specific set of questions.

\item Search for connections through hierarchical relations: the SearchHierarchicalConnections() function. In this method, we look for connections between pairs of concepts located in the same question, using only hierarchical relationships (rdfs:subClassOf and ome:hasChild) as relationships between neighboring concepts. This type of connection is guaranteed to be present between any two concepts in the ontology, but it is not necessarily optimal. We select the shortest connection from the resulting set of this method.

\item Search for connections through any relations: Search All Connections(). In this method, we look for connections between pairs of concepts located in the same question, using all types of relations as relations between neighboring concepts. We select the shortest connection from the resulting set of this method.

\item Comparison of shortest distances with Previous methods: CompareDistances() function. In this method, we check a link using all types of links, which is shorter than the corresponding link using only hierarchical links. If such a connection is found, then the optimal connection between a pair of concepts is confirmed.
Table 3 shows examples of optimal connections found in ontology, indicating the lengths of these connections.
\end{enumerate}
\section{Conclusion}
This article examines the structure of the OntoMathEdu ecosystem, designed to support the process of personalized teaching of school mathematics, and also provides an algorithm for enriching ontology, based on the contextual proximity of concepts in geometry tasks, which contributes to improving the horizontal connectivity of ontology.

Future work is related to the implementation of the personalization component in all implemented test generators, as well as the overall development of these components. In addition, it is planned to develop a component for the automatic analysis of a detailed response by drawing up a framework (scheme) of the solution, and filtering tasks by types of these frameworks.

\begin{table*}
  \caption{Examples of optimal connections found}
  \label{tab:chains}
  \begin{tabular}{p{0.45\linewidth}p{0.15\linewidth}p{0.15\linewidth}p{0.15\linewidth}}
    \toprule
    Source theorem & First concept & Second concept & Link length \\
    \midrule

\textbf{If the diagonals of a given quadrilateral are perpendicular, then this quadrilateral has:
1) equal middle lines,
2)<perpendicular> <middle lines>
3) equal opposite angles} & Perpendicular & Triangle middle line & 3 \\
 &  &  &   \\

\textbf{If a circle can be described around a quadrilateral, then
1) the center of the circle is equidistant from the sides of the quadrilateral 2) the bisectors of the angles of the quadrilateral intersect at one point
3) the sum of <opposite angles> is equal to <right angle>
} & Opposite corners of a quadrilateral & Right angle & 4 \\
 &  &  &   \\

\textbf{<Segments> connecting the midpoints of the sides of any quadrilateral and the segment connecting the midpoints of <diagonals>} & The middle of the segment & Diagonal of the polygon & 3 \\
 &  &  &   \\

\textbf{The sum of two axial symmetries with intersecting axes can be replaced by <rotation> around the <point> at which the ax-es intersect at an angle equal to the angle between the axes} & Point & Rotation angle & 4 \\
 &  &  &   \\

\bottomrule
\end{tabular}
\end{table*}

\begin{acknowledgments}
The article was prepared within the framework of the government task of the Federal State Institution of the Federal Research Center NIISI RAS for 2022-2024 (topic No. 0580-2022-0014 (FNEF-2022-0014)).
\end{acknowledgments}

%%
%% Define the bibliography file to be used
\bibliography{sample-ceur}

\appendix

\end{document}